\documentclass[11pt]{article}

\usepackage{acl}
\usepackage{times}
\usepackage{latexsym}
\usepackage[T1]{fontenc}
\usepackage[utf8]{inputenc}
\usepackage{microtype}
\usepackage{inconsolata}

\usepackage{amsmath}
\usepackage{amssymb}
\usepackage{pifont}
\usepackage{multirow}
\usepackage{multicol}
\usepackage{booktabs}
\usepackage{graphicx}
\usepackage{threeparttable}
\usepackage{enumitem}

\newcommand{\cmark}{\ding{51}}%
\newcommand{\xmark}{\ding{55}}%

\begin{document}
\title{CSTRL: Context-driven Sequential Transfer Learning for Abstractive Radiology Report Summarization}

\author{\textbf{Mst. Fahmida Sultana Naznin\textsuperscript{1}},
 \textbf{Adnan Ibney Faruq\textsuperscript{1}},
 \textbf{Mostafa Rifat Tazwar\textsuperscript{1}},
\\
 \textbf{Md Jobayer\textsuperscript{2}},
 \textbf{Md. Mehedi Hasan Shawon\textsuperscript{2}},
 \textbf{Md Rakibul Hasan\textsuperscript{2,3}}
\\
\\
 \textsuperscript{1}Bangladesh University of Engineering and Technology, Dhaka 1000, Bangladesh
\\
 \textsuperscript{2}BRAC University, Dhaka 1212, Bangladesh
\\
 \textsuperscript{3}Curtin University, Bentley WA 6102, Australia
\\
 \small{
   \textbf{Correspondence:} \href{mailto:rakibul.hasan@curtin.edu.au}{rakibul.hasan@curtin.edu.au}
 }
}

\maketitle

\begin{abstract}
A radiology report comprises several sections, including the Findings and Impression of the diagnosis. Automatically generating the Impression from the Findings is crucial for reducing radiologists' workload and improving diagnostic accuracy. Pretrained models that excel in common abstractive summarization problems encounter challenges when applied to specialized medical domains, largely due to the complex terminology and the necessity for accurate clinical context. Such tasks in medical domains demand extracting core information, avoiding context shifts, and maintaining proper flow. Misuse of medical terms can lead to drastic clinical errors. To address these issues, we introduce a sequential transfer learning that ensures key content extraction and coherent summarization. Sequential transfer learning often faces challenges like initial parameter decay and knowledge loss, which we resolve with the Fisher matrix regularization. Using MIMIC-CXR and Open-I datasets, our model, \textit{CSTRL}---\textbf{C}ontext-driven  \textbf{S}equential \textbf{TR}ansfer \textbf{L}earning---achieved state-of-the-art performance: 56.2\% improvement in BLEU-1, 40.5\% in BLEU-2, 84.3\% in BLEU-3, 28.9\% in ROUGE-1, 41.0\% in ROUGE-2 and 26.5\% in ROGUE-3 score over benchmark methods. We further analyze factual consistency scores while preserving the medical context. Our code is publicly available at \url{https://github.com/fahmidahossain/Report_Summarization}.
\end{abstract}

\section{Introduction}
A radiology report summarizes a radiologist’s analysis of imaging data, detailing sections like Type of Exam, Clinical History, Technique, Findings, and Impression \cite{1Cai2023}. Findings are observations from the examination of diagnosed body parts, classified as normal, abnormal, or potentially abnormal \cite{43Fazeela2024}. Impression is a summary of Findings, including possible causes \cite{43Fazeela2024}. Quality of contextual, informative, and factual correctness of generated Impression is subpar due to limited training for writing Impression ($\leq$1 hour/year for 86\% of radiologists) \cite{2Hartung2020}. Synthesizing Impression from the Findings is, therefore, crucial in automated radiology report summarization.

Writing Impression from Findings falls under the \textit{abstractive summarization} task that distills key clinical insights from diagnostic data \cite{3Herts2021}. Extractive summarization selects sentences from the original text to create a summary, while abstractive summarization generates a new paragraph to summarize the content of the original document \cite{1Cai2023}. Abstractive summarization is more complex than extractive summarization but yields more flexible and concise summaries \cite{1Cai2023}. However, systematic evaluation of abstractive summarization across diverse domains is limited \cite{32zhang2020pegasus}. One challenge of abstractive summarization in the clinical domain is managing the balance between providing concise, clinically relevant information and avoiding excessive technical language. For example, \textit{“Liver metastases have enlarged”} is more straightforward and actionable than \textit{“There is redemonstration of multiple liver lesions consistent with metastases, which have increased in size in the interval”} \cite{2Hartung2020}. Including irrelevant Findings can create ambiguity. For example, \textit{“diverticulosis without diverticulitis”} \cite{2Hartung2020}: diverticulosis indicates the presence of diverticula in the colon, which is typically benign, but, without proper context, it may raise concerns about potential complications and suggest that diverticulitis (inflammation of diverticula) could develop. To prioritize core sentences of Findings and preserve context, we propose a novel approach combining Sequential Transfer Learning with a Knowledge Distillation (KD) strategy, which automates the creation of Impression from the Findings of a radiology report.

Deep learning techniques for abstractive summarization face specific challenges with radiology corpora. A key limitation is the difficulty in evaluating performance on datasets dense with radiology-specific terminology, along with significant differences in word distributions between general and radiology report domains \cite{1Cai2023}. Previous works (\autoref{sec_rl}) have several shortcomings. Firstly, there is a lack of a proper methodology for generating Impression that focuses on primary observations \cite{2Hartung2020}. We propose an optimized Gap Sentence Generation (GSG) technique to identify key Findings. Secondly, subtle differences in terminology can obscure the context and misrepresent the clinical scenario \cite{41codish_model_ambiguity_2024}. In response, we propose a contextual tagging method. Thirdly, poor BLEU scores of recent studies indicate that the generated Impression section often fails to accurately reflect core points from the Findings section \cite{1Cai2023}. It affects the readability, informativeness, and reliability of reports for clinical use. Lastly, reducing dimensionality and computational complexity for real-time production is a challenge \cite{57-10.1145/3503181.3503211, 58che2015distilling, 59liu2020noisy}, for which we propose a KD method. Our key contributions in constructing Impression through proper clinical methodology with high performance and factual consistency are as follows:
\begin{itemize}[nosep]
\item We propose an optimized GSG technique for pivotal sentence identification.
\item We introduce a sequential transfer learning framework with KD using the T5 (Text-to-Text Transfer Transformer) model, fine-tuned on a radiology corpus. This approach aims to transfer knowledge from the GSG task to summarization effectively and gradually. To mitigate catastrophic forgetting during this sequential knowledge transfer, our model includes Fisher matrix regularization with penalty adjustment.
\item We introduce a contextual tagging approach with a Named Entity Recognition (NER) system to capture and preserve essential contextual information by entity linking to the MRCONSO database, a component of the Unified Medical Language System (UMLS) Metathesaurus. It provides a structured representation of medical concepts and terms from various sources.
\item We compare our proposed summarization model, CSTRL, against state-of-the-art models trained on the same dataset. We further train decoder-only models for comparison. CSTRL consistently outperforms all baselines across various settings. We further analyze factual consistency and conduct a comprehensive human evaluation to validate the quality and reliability of the generated summaries.
\end{itemize}

\begin{figure*}[t!]
    \centering
    \includegraphics[width=\textwidth]{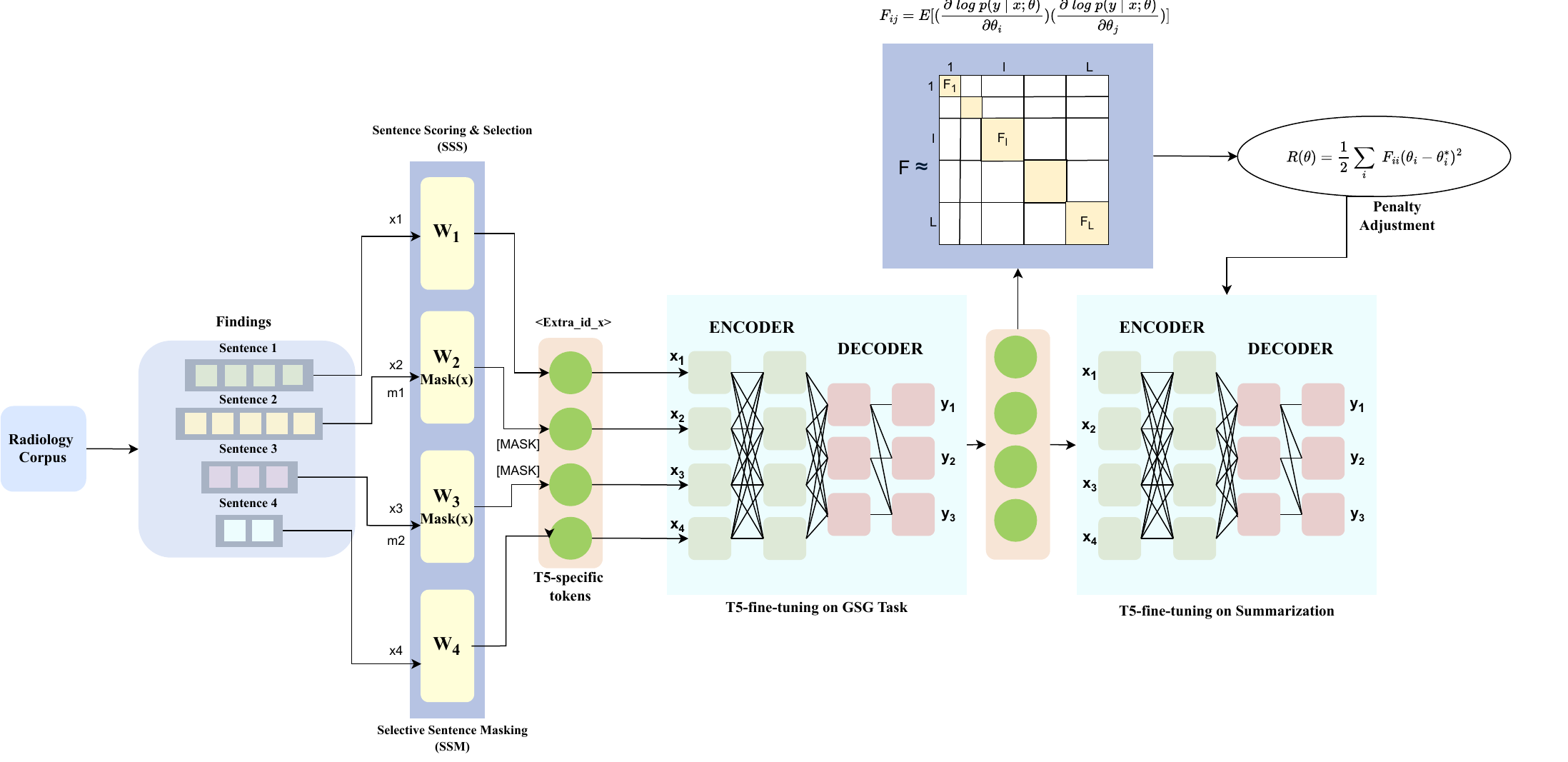}
    \caption{Workflow of the proposed scheme for Sequential Transfer Learning. Firstly, sentences from the radiology corpus are scored with ROUGE and BLEU metrics, and top-scoring sentences are masked. Secondly, the T5 model is trained on the GSG task to predict the masked sentences. The Fisher Information matrix is calculated to identify critical parameters. Lastly, the penalty term is adjusted during fine-tuning for summarization.}
    \label{fig_gsg_mask}
\end{figure*}
\section{Related Work}\label{sec_rl}

In recent years, deep learning has significantly improved neural abstractive summarization tasks in NLP \cite{5rush2015neural, 6chopra2016abstractive, 7tan2017abstractive, 8zhou2017selective, 9li2017deep}. Traditional models are primarily trained on general datasets like CNN/Daily Mail \cite{10nallapati2016abstractive} and Gigaword Corpus \cite{11sutskever2014sequence}. The attention-based seq2seq model by \citet{5rush2015neural} and its extension by \citet{6chopra2016abstractive} laid the groundwork for these advancements. Recently, large-scale pre-trained language models have shown impressive summarization results \cite{13karn2021few, 14raffel2020exploring, 15kieuvongngam2020automatic, 16chang2021jointly}. \citet{15kieuvongngam2020automatic} employed BERT and GPT for summarizing COVID-19 research. However, unique word distributions in radiology corpora limit the applicability of these techniques \cite{1Cai2023}.

\citet{20zhang2018learning} first explored automatic radiology Impression generation, followed by \citet{21macavaney2019ontology}, who introduced an ontology-aware pointer-generator model to enhance summarization quality. A background-augmented pointer-generator network with copy and background-guided decoding was proposed in \citet{added1zhang-etal-2020-optimizing}, and a word graph captured critical words and relations in \citet{added2hu-etal-2021-word}. Anatomies were extracted, radiographs encoded, and fused with anatomy-enhanced co-attention in \citet{added3hu-etal-2023-improving-radiology}. \citet{added4sotudeh-gharebagh-etal-2020-attend} enhanced clinical summarization by augmenting ontological terms, and \citet{added5hu-etal-2022-graph} integrated Findings by a unified framework with knowledge via text and graph encoders. \citet{added6karn-etal-2022-differentiable} proposed a two-step extractive-abstractive method using a Differentiable Multi-Agent Actor-Critic framework.

\citet{22li2018hybrid} proposed a hybrid retrieval-generation agent that integrates human knowledge with neural networks for medical report generation. Models like OpenAI GPT \cite{16chang2021jointly}, BERT \cite{25devlin2019bert}, ELMo \cite{26peters2018deep}, and XLNet \cite{27yang2019xlnet} have improved various NLP tasks through extensive external knowledge. BioBERT \cite{28lee2020biobert} captures semantic features in the biomedical field, pre-trained on large corpora like PubMed abstracts and PMC articles. Recently, \citet{1Cai2023} proposed a pre-trained language model, ChestXrayBert, specifically designed for summarizing chest radiology reports. However, these studies may struggle with effective Impression generation methodologies.

\section{Method}
We adopt a sequential two-step approach to fine-tune the T5 model \cite{14raffel2020exploring} using a dataset on radiology corpus optimized by the GSG methodology \cite{32zhang2020pegasus}. In the first step, the model is trained on a GSG task, where it predicts masked sentences, thereby gaining a deeper understanding of the crucial statements for crafting coherent summaries in the Impression section. In the second step, we fine-tune the model for summarization using the learned weights as initial parameters and employ Fisher matrix regularization for effective GSG-trained knowledge transfer to the summarization task. Then, the GSG fine-tuned model drives sequential knowledge distillation to achieve dimensionality reduction.

\subsection{Proposed Sequential Transfer Learning}

\paragraph{Optimized GSG Technique.}

We propose a novel pre-training methodology for GSG that extends BERT's Masked Language Modeling (MLM) \cite{1Cai2023} objective by integrating an enhanced version of the GSG strategy from PEGASUS \cite{32zhang2020pegasus}. This approach uses sentence scoring and selection for identification of vital sentences.

\paragraph{Sentence Scoring \& Selection (SSS).}
Unlike PEGASUS, which relies solely on ROUGE, our optimized model employs a composite metric that combines ROUGE and BLEU for a more comprehensive assessment of fundamentals by quantifying the overlap and similarity between sentences. In the radiology domain, consistent terminology and phrasing is used in identical medical conditions \cite{63khorasani2003terminology, 64panicek2016sure, 62centers2016everyday}. This consistency makes n-grams essential for capturing the precise repetition of critical terms to ensure accurate contextual representation \cite{62centers2016everyday}. BLEU's emphasis on exact n-gram matches ensures clinical accuracy \cite{61ibrahim2023systematic, 60santhosh2023understanding}. For each sentence $x_i$ in the document $D$, the priority score $W_i$ is computed as:
\begin{equation}
\begin{split}
W_i = F1(\text{ROUGE}(x_i, D \setminus \{x_i\}) + \\ \text{BLEU}(x_i, D \setminus \{x_i\}))
\end{split}
\end{equation}

\paragraph{Selective Sentence Masking (SSM).}

We incorporate a Selective Sentence Masking (SSM) specifically designed for a radiology corpus. We applied three masking rules: (1) sentences $\geq$ 5 words received 3 masks, (2) sentences = 4 received 2 masks, and (3) sentences $\leq$ 3 words received 1 mask. The set of sentences selected for masking is denoted as \( M = \{ m_1, m_2, \ldots, m_k \} \), derived from a masking function \( \text{Mask}(x) \) based on SSS. Our dataset's sentences are predominantly up to 9 sentences long, with 95\% fitting this length. To create the masked text in the masking function, we replace each sentence in the set \( M \) with a placeholder token \( [MASK] \) in the original text \( x \). By applying this methodology to the dataset, we create a new dataset where selected sentences are systematically masked. One additional column contains the original sentences with the masked portions, while another column specifically lists the masked sentences alongside the existing columns.

\paragraph{Fine-Tuning for GSG Task.}
We replace the [MASK] in the GSG-optimized dataset with T5-specific placeholders (<extra\_id\_x>). The dataset $ D = \{(x_1, y_1), (x_2, y_2), \ldots, (x_n, y_n)\} $ comprises input sequences \(x_i\) and their corresponding target sequences \(y_i\) which are derived from additional columns in the GSG-optimized dataset. Both \(x_i\) and \(y_i\) are tokenized with the pre-trained T5 tokenizer, applying truncation and padding for uniform length. The model is trained to predict the masked sentences.

\begin{figure*}[t!]
    \includegraphics[width=0.88\textwidth]{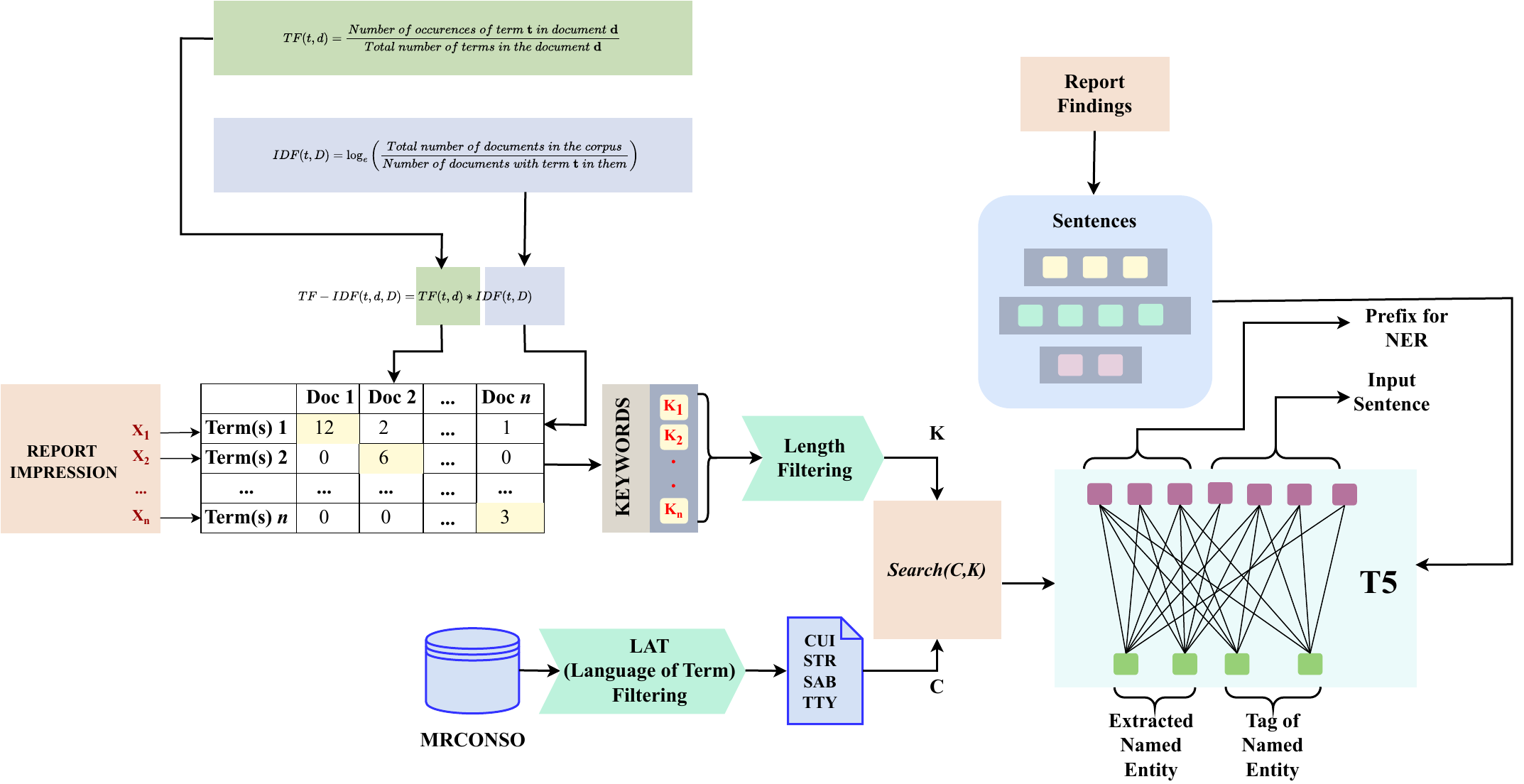}
    \caption{Workflow of the proposed scheme for contextual tagging. Firstly, keywords from the Impression section are extracted using TF-IDF vectorization. Secondly, relevant clinical terms are retrieved from the MRCONSO table, and tags are selected by searching keywords in the filtered table. Finally, the T5 model is trained to predict tags from the Findings section.}
    \label{fig:ner_diagram}
\end{figure*}

\subsection{Sequential Transfer Learning for Summarization}

After fine-tuning the model on the GSG task, we utilize its learned weights as the initial parameters for the summarization task. However, we observe that direct sequential fine-tuning significantly updates the model parameters, which risks losing the valuable knowledge acquired during the GSG training. Our objective is to gradually transfer the knowledge from the GSG-trained model to enhance summarization without overwriting it. To resolve this issue, we implement Fisher Matrix Regularization (FMR), inspired by Elastic Weight Consolidation \cite{38kirkpatrick2017overcoming} to diminish the problem of catastrophic forgetting. When a model erases prior knowledge while assimilating new tasks, compromising its ability to retain past information, it experiences catastrophic forgetting—a fundamental challenge in sequential learning. Deep neural networks, inherently vulnerable to this, require explicit regularization to mitigate its effects. The FMR is widely used as a pivotal mechanism to address this \cite{50-9933427, 51-michel2019regularizing, 52-gupta2021addressing, 53-kutalev2020natural, 54-liu2018rotate, 55-ritter2018online, 56-kemker2018measuring}
We compute the Fisher Information Matrix $F$ (FIM) based on the GSG task to identify key parameters that should be preserved. The FIM is defined as:
\begin{equation}
\begin{split}
F_{ij} = \mathbb{E} \left[ \left( \frac{\partial \log p(y|x; \theta)}{\partial \theta_i} \right) \left( \frac{\partial \log p(y\vert x; \theta)}{\partial \theta_j} \right) \right]
\end{split}
\end{equation}
where \( p (y|x; \theta) \) represents the model's likelihood given the input data \( x \) and parameters \( \theta \). The matrix \( F \) quantifies how sensitive the performance is to changes in each parameter, highlighting those that are paramount for retaining the knowledge gained from the GSG task. During the fine-tuning process on the summarization task, we introduce a penalty term \( R(\theta) \) to limit significant changes to these crucial parameters: \( R(\theta) = \frac{1}{2} \sum_i F_{ii} (\theta_i - \theta^*_i)^2 \)
where \( \theta^* \) denotes the optimal parameters obtained from the GSG task. The penalty term \( R(\theta) \) is dynamically adjusted and gradually reduced over the training epochs. \autoref{fig_gsg_mask} represents the overall architecture. This approach enables effective adaptation to summarization while retaining key knowledge from GSG training.

\subsection{Contextual Tagging for Clinical Precision}

We note that overgeneralization or overcomplication in abstractive summarization leads to significant misinterpretations. Such distortions, especially in clinical contexts, critically alter the intended meaning of Findings. To address this issue, we propose a methodology for developing an optimized NER system to obtain core contextual knowledge. We identify key medical terms from Impression for entity recognition using the Term Frequency-Inverse Document Frequency (TF-IDF) method \cite{44Christian2023}. After applying TF-IDF vectorization, we generate a matrix representing each document in a high-dimensional space. The top \( n \) keywords, denoted as \( K \), are extracted based on TF-IDF scores. Then, we filter data from the MRCONSO table to include only relevant English terms. The table maps medical terminologies by assigning each concept a unique CUI. It preserves context by linking terms to their source vocabulary and language. These identifiers ensure that variations in spelling or phrasing are consistently mapped to the correct concept. It maintains semantic integrity across systems. The filtered dataset from MRCONSO includes the Concept Unique Identifier (CUI), String (STR), Source Abbreviation (SAB), and Term Type (TTY). They serve as a reference list of tags \( C \). We search the extracted keywords \( K \) in \( C \) and make a new column of tags to their corresponding Impression. Then, we train the model using T5 to generate tags from Findings. \autoref{fig:ner_diagram} shows the architecture of proposed contextual tagging.

\begin{figure}[!t]
    \centering
\includegraphics[width=\linewidth]{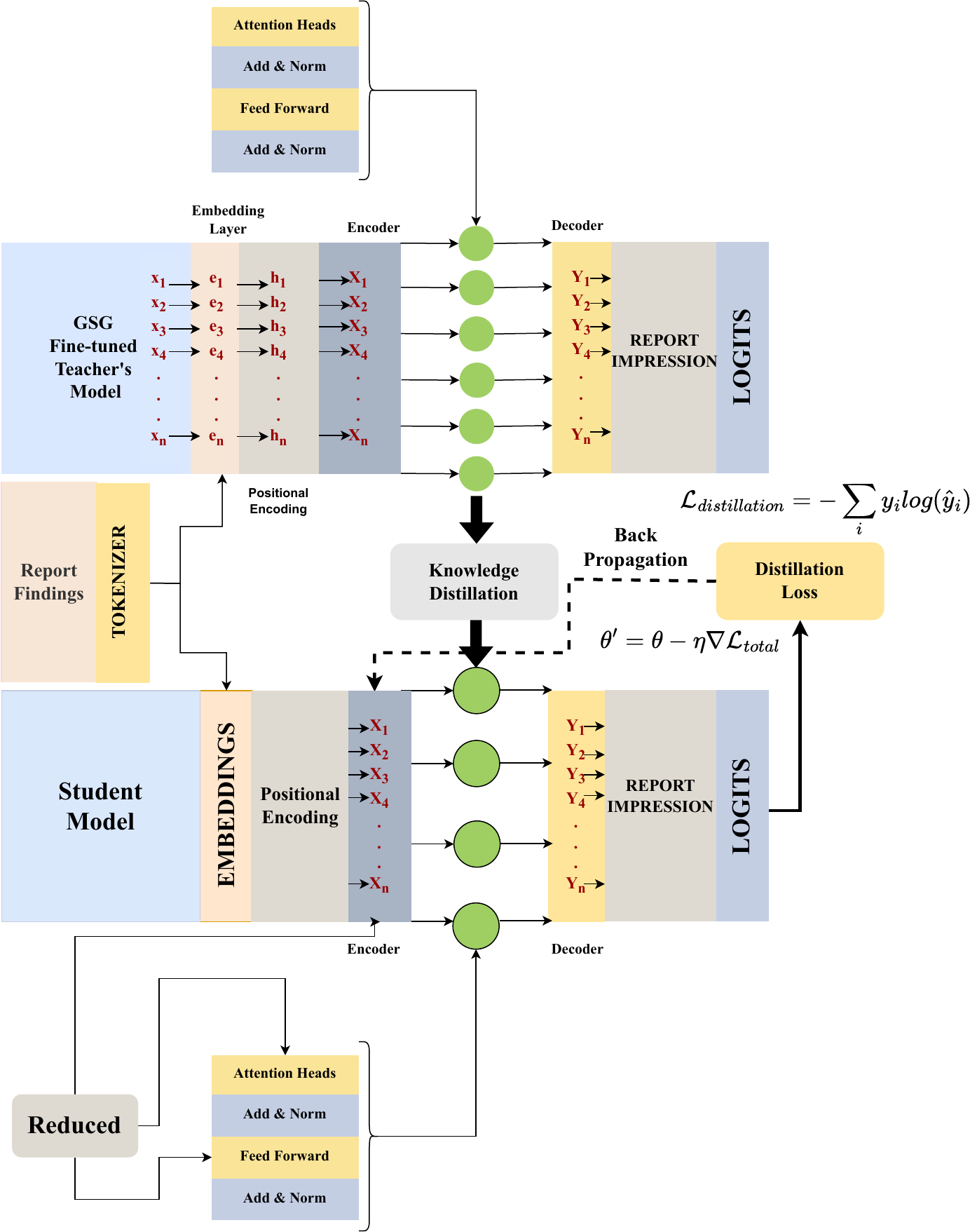}
    \caption{Schematic representation of the proposed knowledge distillation network. The combined loss function is calculated using the reduced feature layers and the Hard Labels loss. Effective back-propagation ensures the optimization of the Student Model.}
    \label{fig_kd_diagram}
\end{figure}

\subsection{Proposed Teacher-Student Framework for Knowledge Distillation}

\begin{figure*}[!t]
    \centering
    \includegraphics[width=0.7\linewidth]{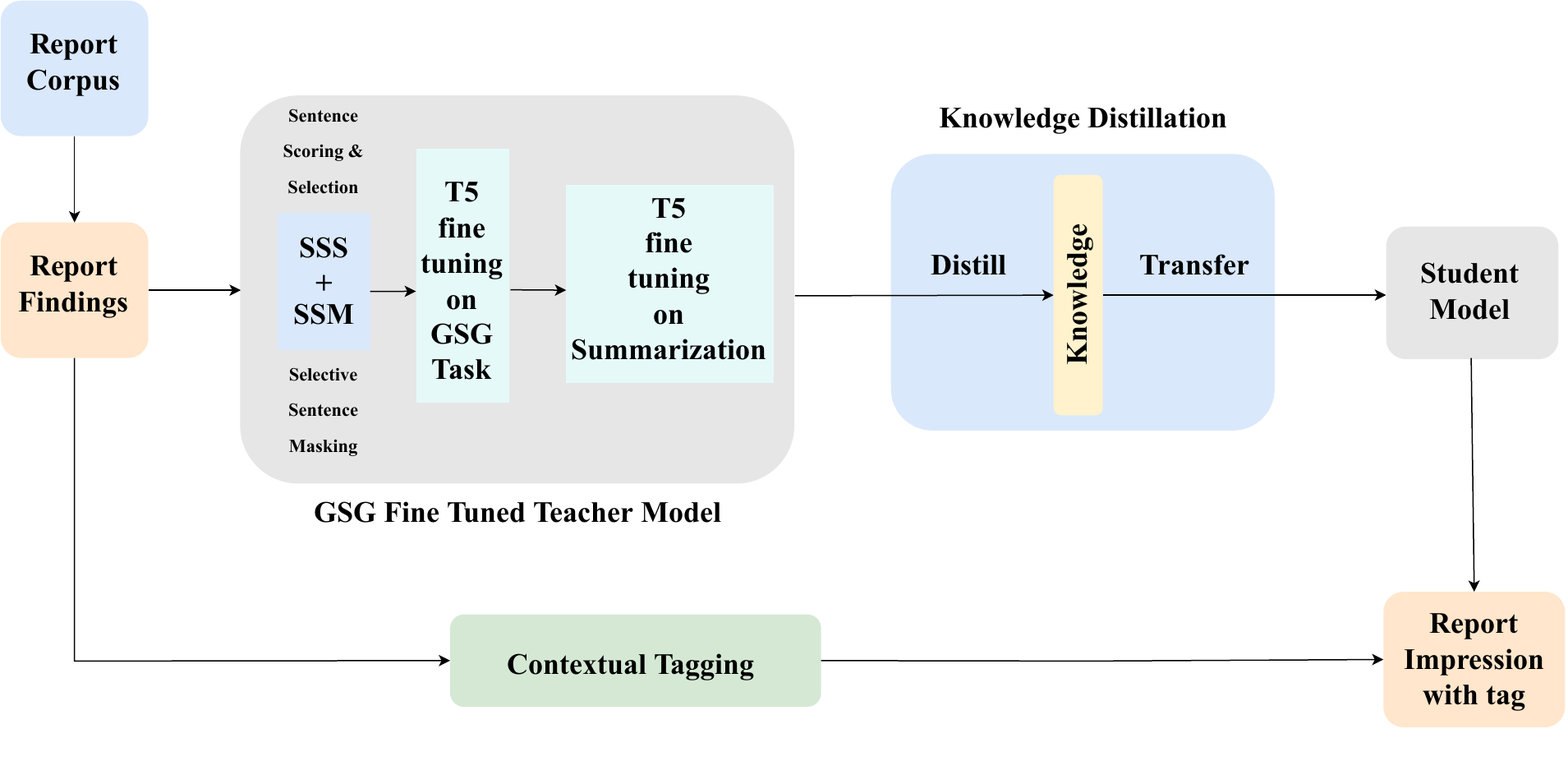}
    \caption{Summary of CSTRL: The GSG Finetuned Teacher model (\autoref{fig_gsg_mask}), using sequential transfer learning, is knowledge distilled, following the knowledge distillation process (\autoref{fig_kd_diagram}). Additionally, a contextual tagging method (\autoref{fig:ner_diagram}) is applied to the generated impressions.}
    \label{fig_summary}
\end{figure*}

The GSG fine-tuned summarization model acts as a teacher, generating logits that serve as soft targets for the student model as shown in \autoref{fig_kd_diagram}. This allows the student to learn from the teacher’s outputs rather than relying solely on hard labels. We employ a sequential training algorithm, where each batch undergoes multiple forward and backward passes. We use the T5 tokenizer to convert Findings and Impressions into token IDs and attention masks. These are fed into an embedding layer, mapping tokens to dense vector representations. The shared embedding framework between teacher and student models ensures consistency, improving performance in downstream tasks.
To train the student model, we combine cross-entropy loss with distillation loss, allowing the student to learn from both ground-truth labels and softened outputs from the teacher. The combined loss is defined as: 

\begin{equation}
\mathcal{L} = (1 - \alpha) \mathcal{L}_{CE} + \alpha \mathcal{L}_{KL}
\end{equation}
where \( \mathcal{L}_{CE} \) is the cross-entropy loss, \( \mathcal{L}_{KL} \) is the Kullback-Leibler (KL) divergence loss between teacher and student logits, and \( \alpha \) is a tunable parameter. The cross-entropy loss \( \mathcal{L}_{CE} \) is given by:
\begin{equation}
\mathcal{L}_{CE} = -\sum_{i=1}^{N} Y_i \log(\text{softmax}(S_i))
\end{equation}

with \( S \) representing student logits, \( Y \) as ground-truth labels, and \( N \) as the number of samples. To compute the distillation loss, we apply temperature scaling to soften the teacher model's logits as follows:
\begin{equation}
\mathcal{L}_{KL} = {T^2} \cdot \text{KLDiv}(\text{softmax}(S/T), \text{softmax}(T_t/T))
\end{equation}
where \( T_t \) refers to teacher logits. \( T \) is the temperature. This softening enables the student to approximate the teacher's output distribution more effectively, and the \( T^2 \) term ensures appropriate gradient scaling during backpropagation. Model parameters are updated via backpropagation to minimize the combined loss. \autoref{fig_summary} summarizes our CSTRL, which uses sequential transfer learning, knowledge distillation, and contextual tagging for generated impressions.

\section{Experiment and Evaluation}
\begin{table}[t!]
\resizebox{\columnwidth}{!}{%
\begin{threeparttable}
\begin{tabular}{@{}llllllll@{}}
\toprule
\multirow{2}{*}{Dataset} & \multirow{2}{*}{\# Reports} & \multicolumn{3}{l}{Findings (Avg.)} & \multicolumn{3}{l}{Impressions (Avg.)} \\ \cmidrule(l){3-8} 
 &  & \# S & \# W/S & \# Source W & \# S & \# W/S & \# Source W \\ \midrule
MIMIC-CXR & 121,975 & 5.47 & 10.08 & 55.09 & 1.94 & 8.50 & 16.46 \\
Open-I & 3,312 & 4.62 & 8.03 & 37.06 & 1.81 & 5.52 & 9.98 \\ \bottomrule
\end{tabular}%
\begin{tablenotes}
    \item S: Sentence, W: Words
\end{tablenotes}
\end{threeparttable}}
\caption{Summary of dataset statistics of Findings and Impression}
\label{tab:dataset_summary}
\end{table}
\subsection{Dataset Description}

We evaluate our model using two public chest X-ray report datasets: MIMIC-CXR \cite{mimic_cxr} and Open-I \cite{openi}, as shown in \autoref{tab:dataset_summary}. MIMIC-CXR contains 227,835 reports, while Open-I, a public radiography dataset from Indiana University, originally included 7,430 reports. Findings in the MIMIC-CXR dataset exhibit an average of 5.47 sentences per Findings, with each sentence containing approximately 10.08 words. It has an average of 55.09 words per Findings, which suggests this dataset is rich in detail and complexity. Conversely, Open-I has slightly fewer sentences per Findings, averaging 4.62 sentences, with each sentence comprising 8.03 words and a total of 37.06 words. Impression reveals similar trends. MIMIC-CXR has an average of 1.94 sentences per Impression, with an average of 8.50 words per sentence, leading to an average of 16.46 words. In contrast, Open-I averages 1.81 sentences per Impression, with 5.52 words per sentence and 9.98 source words.

To ensure the quality of the data for medical report summarization, we apply several filtering criteria inspired by a GPT-Based Radiology Report Optimization study \cite{31ma2020iterative}. We remove (1) incomplete reports that lack either Findings or Impression, (2) reports with a Findings section of fewer than 10 words, and (3) reports with an Impression section of fewer than 2 words. We retain 121,975 reports from MIMIC-CXR and 3,312 reports from Open-I. We then merge these datasets, a total of 125,287 reports. This merged dataset is divided into training, validation, and test sets using an 8:1:1 split.

\subsection{Experimental Setup}

\paragraph{Evaluation Metrics.}

To evaluate the quality of text summarization, we report F1 scores for ROUGE-1, ROUGE-2, and ROUGE-L (unigram, bigram, and longest common subsequence) \cite{46Lin2004}. We also provide BLEU-1, BLEU-2, and BLEU-3 (unigram, bigram, and trigram) \cite{40papineni2002bleu}. In addition, we measure the factual consistency of the generated summaries using SummaC-ZS and SummaC-Conv \cite{44Christian2023}. We set \texttt{granularity = ``sentence''} and \texttt{model\_name = ``vitc''}.

\begin{table*}[t]
\centering
\resizebox{0.7\textwidth}{!}{%
\begin{tabular}{@{}lllllll@{}}
\toprule
Model & R-1 & R-2 & R-L & B-1 & B-2 & B-3 \\ \midrule
Content Selector \cite{added4sotudeh-gharebagh-etal-2020-attend} & 53.6 & 40.8 & 51.8 & -- & -- & --  \\ 
WGSUM (Trans+GAT) \cite{added2hu-etal-2021-word} & 48.3 & 33.3 & 46.7 & -- & -- & --  \\ 
BASE+GRAPH+CL \cite{added5hu-etal-2022-graph} & 49.1 & 33.7 & 47.1 & -- & -- & --  \\ 
Seq2Seq \cite{1Cai2023}       & 35.1 & 23.7 & 35.4 & 24.3 & 11.5 & 4.1 \\
PGN \cite{1Cai2023}           & 36.1 & 23.9 & 35.7 & 23.9 & 11.3 & 4.6  \\
PGN(Cov) \cite{1Cai2023}      & 36.6 & 24.2 & 37.5 & 27.6 & 13.7 & 5.4  \\
RadSum \cite{1Cai2023}        & 37.9 & 25.5 & 39.2 & 28.0 & 13.9 & 5.5  \\
TransAbs \cite{1Cai2023}      & 37.7 & 26.9 & 38.7 & 28.1 & 14.1 & 5.7  \\
BART \cite{1Cai2023}          & 39.4 & 27.3 & 39.7 & 28.3 & 14.2 & 5.8  \\
ChestXRayBERT \cite{1Cai2023} & 41.3 & 28.6 & 41.5 & 28.5 & 14.4 & 6.1  \\ 
BASE+AP+DCA \cite{added3hu-etal-2023-improving-radiology} & 47.6 & 32.0 & 46.1 & -- & -- & --  \\ 
Meta-Llama-3-8B \cite{addedextrazhao2024improvingexpertradiologyreport} & -- & -- & 29.0 & -- & -- & 9.4 \\
 Meta-Llama-3-1B (Zero-shot) & 0.2 & 0.1 & 0.2 & 0.8 & 0.3 & 0.1 \\
 Meta-Llama-3-1B (Few-shot) & 30.5 & 12.9 & 27.9 & 31.2 & 55.2&67.4 \\
CSTRL (Ours) & \textbf{58.1} & \textbf{48.5} & \textbf{56.5} & \textbf{65.0} & \textbf{47.9} & \textbf{38.9}   \\ \bottomrule
\end{tabular}%
}
\caption{Performance comparison across various models on the combined OPEN-I and MIMIC-CXR test sets.}
\label{tab:model_values}
\end{table*}

\paragraph{Implementations and Model Details.}
We use PyTorch to implement our model on an NVIDIA A100 GPU. We configure the Teacher Model with 6 layers, 512 embedding dimensions, and 8 attention heads, while the Student Model has 3 layers, 128 dimensions, and 4 attention heads. Both use a vocabulary size of 32,128. Inputs are tokenized to 512 tokens, targets to 256. Models are trained for 20 epochs with a batch size of 32, using AdamW with a 0.003 learning rate. Logits are adjusted with a temperature of 20, and distillation loss is scaled by $\alpha$ = 0.7. ROUGE and BLEU with the T5-small are used for key token identification, masking 3 sentences per Findings. We conduct a grid search throughout all experiments for tuning hyperparameters.

\begin{table}[t]
\centering
\resizebox{\columnwidth}{!}{%
\begin{tabular}{@{}ccc|llllll@{}}
\toprule
\multicolumn{3}{l|}{Ablation Settings} & \multirow{2}{*}{R-1} & \multirow{2}{*}{R-2} & \multirow{2}{*}{R-L} & \multirow{2}{*}{B-1} & \multirow{2}{*}{B-2} & \multirow{2}{*}{B-3} \\ \cmidrule(r){1-3}
GSG & Fisher Matrix & Layer Unfreezing &  &  &  &  &  &  \\ \midrule
\xmark & \xmark & \xmark & 55.9 & 45.2 & 54.2 & 63.2 & 45.4 & 35.4 \\
\cmark & \xmark & \xmark & 55.9 & 45.2 & 54.2 & 63.2 & 45.4 & 35.4 \\
\cmark & \cmark & \xmark & 58.2 & 48.5 & 56.5 & 65.0 & 47.9 & 38.9 \\
\cmark & \xmark & \cmark & 53.4 & 43.1 & 51.9 & 61.5 & 42.3 & 32.3 \\ \bottomrule
\end{tabular}%
}
\caption{Ablation study of the effect of varying techniques in the proposed CSTRL architecture.}
\label{tab:training_params}
\end{table}

\begin{table}[t]
\centering
\resizebox{\columnwidth}{!}{%
\begin{tabular}{@{}llllllll@{}}
\toprule
Model & R1 & R-2 & R-L & B-1 &B-2& B-3 \\ \midrule
        CSTRL-T      & 58.1 & 48.5 &56.5 & 65.0 &47.9 & 38.9\\
        CSTRL-S ($\times 8$)     & 49.8  &37.9 & 48.8   & 61.0   & 37.1  & 26.3\\
         CSTRL-S ($\times 16$)  &  47.8  & 36.3   & 46.7  & 58.5   & 35.9   & 25.2   \\
        CSTRL-S ($\times 32$)     & 46.0  &34.9 & 44.9   & 56.4   & 34.6   & 24.3\\
\bottomrule
\end{tabular}%
}
\caption{Performance of reduced models on different distillation schemes.}
\label{tab:teacher_student_models}
\end{table}

\subsection{Results and Discussion}
\begin{figure}[t!]
    \centering
    \includegraphics[width=\columnwidth]{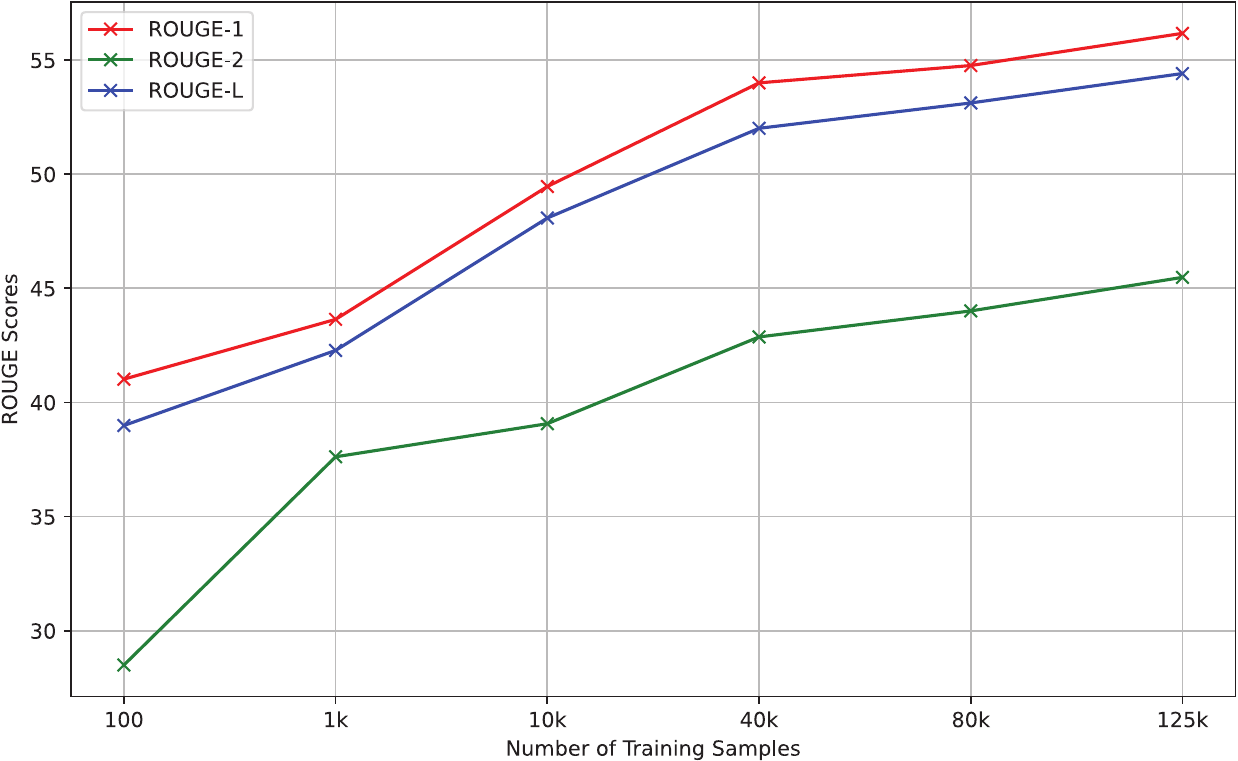}
    \caption{Fine-tuning with limited supervised examples. The solid lines are  CSTRL fine-tuned on different numbers of datasets.}
    \label{fig:dataset_perf}
\end{figure}

We compare our proposed summarization model, CSTRL, with state-of-the-art neural network models trained on the same dataset. We also train a decoder-only Meta-Llama-3-1B model (similar in size to T5's 770M). The results in \autoref{tab:model_values} show that our model outperforms others in all evaluation metrics, including ROUGE-x (R-x) and BLEU-x (B-x). To simulate low-resource summarization, we randomly select subsets of \(10^k\) samples (where \(k = 2, 3, 4, 5\)), including 40,000 and 80,000 samples, to train our model, selecting the best validation checkpoint after convergence. As shown in \autoref{fig:dataset_perf}, fine-tuning with 40,000 examples (32.8\% of the dataset) achieves summarization quality similar to full-data training. As shown in \autoref{tab:model_values}, our model outperforms ChestXRayBERT by 56.2\% in BLEU-1, 40.5\% in BLEU-2, and an extraordinary 84.3\% in BLEU-3. The performance gain in CSTRL comes from two factors: T5 base model and an effective sequential learning strategy. We select T5 due to its exploration of transfer learning with a unified text-to-text transformer. This transformer has a bidirectional encoder and a unidirectional decoder, which distinguishes it from traditional Seq2Seq models. The bidirectional encoder enhances context understanding by analyzing the entire input sequence, enabling better context capture and management of long-range dependencies. We achieve improvements of 28.9\%, 41.0\%, and 26.5\% in ROUGE-1, ROUGE-2, and ROUGE-L, respectively. Fine-tuning on radiology corpora has greatly improved the accuracy of the model. Using GSG with Fisher Matrix Regularization facilitates impactful sequential learning, transitioning from key sentence identification to summarization, as detailed in the ablation study.

\subsection{Performance Evaluation and Ablation Study}

\paragraph{Effect of GSG Fine-Tunning.}
As shown in \autoref{tab:training_params1}, we predict masked sentences and evaluate using ROUGE and BLEU values for different batch sizes. In abstractive summarization, capturing inter-sentence relationships is crucial. We employ GSG, where sentences are masked based on ROUGE and BLUE, akin to an extractive summary. This method effectively accommodates both large and small datasets. Across the three batch sizes, the ROUGE and BLEU scores are quite similar, showing consistent performance across various configurations. The best configuration appears to be a batch size of 32 and a learning rate of 0.003, which provides a BLEU-1 of 55.7, a BLEU-2 of 35.8, and a BLEU-3 of 28.1.

\begin{table}[!t]
\resizebox{\columnwidth}{!}{%
\begin{threeparttable}
\begin{tabular}{@{}llllllll@{}}
\toprule
BS & LR & R-1 & R-2 & R-L & B-1 & B-2 & B-3 \\ \midrule
8 & 0.001 & 39.4 & 25.5 & 35.9 & 54.6 & 34.5 & 26.8 \\
16 & 0.003 & 39.5 & 25.7 & 35.1 & 55.1 & 34.1 & 27.2  \\
\textbf{32}& \textbf{0.003} & \textbf{39.7} & \textbf{26.2} & \textbf{36.3} & \textbf{55.7} & \textbf{35.8} & \textbf{28.1}\\ \bottomrule
\end{tabular}%
\begin{tablenotes}
    \item BS: Batch size, LR: Learning rate
\end{tablenotes}
\end{threeparttable}%
}
\caption{Performance of GSG task with Different Hyperparameters.}
\label{tab:training_params1}
\end{table}

\paragraph{Effect of Fisher Matrix Regularization with Penalty Adjustment.}

We note that the performance metrics show identical values with and without the GSG framework, as shown in \autoref{tab:training_params}, for facing catastrophic forgetting. However, our experiments reveal differences beyond the third decimal point. Direct fine-tuning on the summarization task results in substantial updates to the model parameters, potentially compromising the knowledge gained during GSG training. After applying Fisher matrix regularization, we observe significant improvements in the results. These enhancements are due to the gradual adjustment of key parameters encapsulated within the Fisher matrix. As a result, ROUGE-1, ROUGE-2, ROUGE-L, BLUE-1, BLUE-2 and BLUE-3 scores exhibit a notable 4.1\%, 7.3\%, 4.2\%, 2.8\%, 5.5\% and 9.9\% improvements, respectively. We explore an alternative to Fisher matrix regularization by implementing a gradual layer unfreezing strategy. Initially, we freeze certain layers to retain knowledge from the previous fine-tuned task (GSG). Then, we slowly unfreeze these layers to minimize the risk of losing critical information. This approach reduces performance, decreasing ROUGE-1, ROUGE-2, ROUGE-L, BLEU-1, BLEU-2, and BLEU-3 scores by 4.5\%, 4.6\%, 4.2\%, 2.7\%, 6.8\%, and 8.7\%, respectively (\autoref{tab:training_params}).

\begin{table*}[t!]
\centering
\resizebox{0.7\textwidth}{!}{%
\begin{tabular}{lccccc}
\toprule
\textbf{Model} & \textbf{Readability} & \textbf{Factual Correctness} & \textbf{Informativeness} & \textbf{Redundancy} & \textbf{Completeness} \\ \midrule
CSTRL-T              & 4.62 & 4.39 & 4.31 & 4.11 & 4.75 \\
CSTRL-S ($\times$8)  & 4.59 & 4.29 & 4.28 & 4.73 & 4.54 \\ \bottomrule
\end{tabular}%
}
\caption{Human evaluation results based on summarization quality. Higher scores indicate better performance.}
\label{tab:model_comparison_1}
\end{table*}

\paragraph{Effect of the Distillation Operation on Computation.}

As shown in \autoref{tab:teacher_student_models}, we scale the network parameters to investigate the CSTRL Student model's (CSTRL-S) performance compared to the CSTRL Teacher model (CSTRL-T). Reducing the parameter size significantly drops the performance from 20.8\% to 15.3\%, 28.04\% to 21.9\%, 20.5\% to 13.6\%, 13.2\% to 6\%, 27.8\% to 22.5\%, and 37.5\% to 32.39\% in ROUGE-1, ROUGE-2, ROUGE-L, BLEU-1, BLEU-2, and BLEU-3, respectively. In \autoref{tab:model_comparison_2}, we measure complexity and computational efficiency. Inference time is reduced by up to 63.1\% compared to CSTRL-T. We compare our model, CSTRL, with different settings against widely used baselines. We observe that CSTRL models (T and S variants) are highly efficient. They outperform existing models by reducing computational demands (GFLOPS, GMACs) while maintaining compact parameter sizes and competitive performance. The CSTRL-S variants (×8, ×16, ×32) exhibit a scalable efficiency trade-off: as the model scales down (8× to 32×), computational load decreases, achieving 2.5 to 8 times faster inference than ClinicalBERT or BART and 7 to 20 times faster than XrayGPT, despite using 10 to 100 times fewer parameters. In contrast, larger models like Meta-Llama-3-1B and XrayGPT demand significantly higher computational resources—XrayGPT requires 45 times more GMACs than CSTRL-S 32×—and have considerably slower inference (4s vs. 0.19s). We achieve near-real-time performance with minimal resources, bridging lightweight deployment and clinical-grade AI.

\subsection{Factual Consistency}
We analyze how sequential learning strategies affect the factual consistency of CSTRL using two key models. We evaluate the alignment of the generated summaries with the original documents in the test set. We confirm that the samples used as train data are not present in the validation part. As shown in \autoref{tab:Fact_compare}, the model trained with the GSG Framework and Fisher matrix regularization produces the most consistent Impression of Findings. Sequential learning enhances this consistency. For SummaC-Conv, the variance between Impressions in the original report and those generated without sequential learning (W/O STRL) is 0.020, corresponding to 5.54\%. After implementing sequential learning (STRL), this variance decreases to 0.014, representing 3.87\%. This change indicates a performance enhancement of 1.67\%. In the case of SummaC-ZS, the GSG implementation achieves a variance of 0.002 (3.13\%). After adjusting for sequential learning, the score was 0.028, reflecting 6.25\%. This results in an improvement of 3.12\%. We deem reduced variance between the original and predicted impressions indicative of a loss of nuance. This technique eliminates contextual errors caused by subtle changes.

\subsection{Human Evaluation}

We conduct expert evaluations to correlate ROUGE and BLEU improvements with human judgments. We randomly selected 50 samples from the dataset. We compare summaries generated by our teacher model against its distilled students. Five volunteers participated, including three radiology researchers and two clinical radiologists. They rate the samples on a scale from 1 (very poor) to 5 (very good). \autoref{tab:model_comparison_1} shows the close performance of both models. Upon inspection, we find that student summaries are rather telegraphic in terms of factual correctness.

\begin{table}[t]
\centering
\resizebox{\columnwidth}{!}{%
\begin{tabular}{@{}lllllll@{}}
\toprule
 
Model & GFLOPS & GMACs & Params & Inference Time\\ \midrule
CSTRL-T    & 71.7 & 35.8  & 60.50M&0.521s  \\
 CSTRL-S ($\times 8$)   & 17.9& 9.0 & 11.20M&0.330s
  \\
CSTRL-S ($\times 16$)    & 10.6 & 5.3 & 6.40M&0.220s\\
CSTRL-S ($\times 32$) & \textbf{5.6} & \textbf{2.8}& \textbf{3.63M}& \textbf{0.192s}\\
Meta-Llama-3-1B & 1270 & 632.7 &1.24B& 1.527s\\
ClinicalBERT \cite{clinicalbertmodelingclinicalnotes} & 50 & 40 & 110M& 1s
\\
BART \cite{bartdenoisingsequencetosequencepretraining} & 100 & 200 & 140M& 2s\\
XrayGPT \cite{xraygpt} & 250 & 500 & 350M& 4s\\

\bottomrule
\end{tabular}%
}
\caption{Comparison of models with original and distilled versions based on FLOPs, MACs, and Parameters.}
\label{tab:model_comparison_2}
\end{table}

\begin{table}[!tb]
    \centering
    \begin{tabular}{@{}lll@{}}
        \toprule
        Summary Type & SummaC-ZS & SummaC-Conv \\ \midrule
        Original       & 0.361         & 0.064        \\
        STRL       & 0.341         & 0.068         \\
        W/O STRL        & 0.375         & 0.066         \\
        \bottomrule
    \end{tabular}
    \caption{Performance comparison of factual consistency.}
    \label{tab:Fact_compare}
\end{table}

\section{Conclusion}

This paper proposes CSTRL---Context-driven Sequential Transfer Learning---a novel fine-tuning framework for generating factually correct and coherent radiology report summaries. CSTRL follows a dual-stage process, where it is fine-tuned on both the GSG and summarization tasks. We show in the ablation study that it not only improves summarization quality, as measured by ROUGE and BLEU scores, but also enhances factual correctness. To the best of our knowledge, we first introduce a fusion of methodologies for generating accurate impressions and preserving context within the clinical domain.

\section*{Limitations}
Our proposed model is evaluated on two datasets, both containing predominantly short sentences in the Impression section. A key limitation is that the model's predicted Impressions are similarly concise, likely due to the bias in the training data. As a result, the model may encounter difficulties in generating longer, more detailed impressions.

\section*{Acknowledgement}
This work was supported by resources provided by the Pawsey Supercomputing Research Centre with funding from the Australian Government and the Government of Western Australia. We also gratefully acknowledge the valuable contributions of the five volunteers who dedicated their time and expertise to evaluate our model outputs.

\bibliography{references}

\end{document}